\documentclass[11pt]{article}
\usepackage[a4paper,margin=2.3cm]{geometry}
\usepackage[T1]{fontenc}
\usepackage{lmodern}
\usepackage{microtype}
\usepackage{booktabs}
\usepackage{graphicx}
\usepackage{float}
\usepackage{caption}
\usepackage{url}
\usepackage{xcolor}
\usepackage[colorlinks=true,linkcolor=black,citecolor=black,urlcolor=blue!50!black]{hyperref}
\begin{document}

\title{Embeddings Are Overpriced? A Frozen 16 MB Vocabulary for Small and Ternary Language Models}
\author{\textbf{Sai Kiran Bathula} \\ QLNI, New South Wales, Australia \\ \href{mailto:saikiranbathula@qlni.io}{saikiranbathula@qlni.io}}
\date{}
\maketitle

\begin{abstract}
In a small language model the vocabulary is the single largest part. In Gemma 3 270M the embedding table holds 168M of 268M parameters, and because quantisation methods leave the embedding in 16-bit, it is also most of the file once the rest of the model is compressed. We ask what that table actually buys. We replace it with one frozen table of 512-bit codes, 16 MB, with no trainable parameters, derived once from a pretrained embedding by principal-component projection and sign binarisation, and we train the same Gemma-shaped body from scratch on the same 2.05B tokens with and without it. The learned table buys 0.20 nats of validation loss for 2.6 times the trainable parameters, but on LAMBADA, HellaSwag and PIQA the frozen-table model is better or tied; only ARC-Easy favours the learned table. With a ternary BitNet b1.58 body the whole model is a 46 MB file, still ties the float model on LAMBADA and PIQA, and on a laptop CPU generates 615 to 740 tokens per second in 62 MB of memory, five to six times the speed of the float model's 8-bit GGUF in llama.cpp at a seventh of the memory. The frozen table wins by 1.7 to 2.3 nats on rare words, accepts 10,000 new words after training with a 3\% cost where the learned table collapses by 43\%, and all of it is released with the weights, the data and every evaluation answer.
\end{abstract}

\begin{center}\small Code, weights of every arm (including intermediate checkpoints), training data, logs and per-question evaluation results: \url{huggingface.co/QLNI/embeddings-are-overpriced}. Preprint DOI: doi.org/10.5281/zenodo.23382634. Released for non-commercial research.\end{center}

\section{Why the vocabulary}

Three things happened to small language models in the last few years. BitNet showed that a transformer can be trained with weights in \{-1, 0, +1\} and keep most of its quality \cite{r1,r2}. Models under a billion parameters became good enough to be useful: Gemma 3 270M, SmolLM, Qwen's smallest sizes. And people started running these models on phones and laptop CPUs, where every megabyte and every byte read per token costs time.

The vocabulary did not get the same attention. Every one of these models carries a learned embedding table: one trained vector per token, usually tied to the output head. For a large model the table is a rounding error. For a small model it is the model. Gemma 3 270M has 268M parameters; 168M of them, 63\%, are the table (Table 1). The feed-forward layers, where the model does most of its computing, are 71M.

Quantisation makes this worse, not better. BitNet keeps its embedding and head in 16-bit. GPTQ and AWQ leave \texttt{embed\_tokens} and \texttt{lm\_head} untouched. llama.cpp stores the output head at 6 to 8 bits. So after the body is compressed, the table is what is left. A Gemma 3 270M with a ternary body and a 16-bit vocabulary would be about 20 MB of body and 335 MB of vocabulary. Microsoft's own BitNet b1.58 2B4T report gives its memory as ``0.4 GB (non-embedding)'', leaving the table out of the headline number \cite{r3}. Everyone accepted this. The table is treated as a fixed cost of having a vocabulary.

We did not want to accept it, because we had a reason to think the table is overpriced. A token embedding has two jobs: tell the body which token came in, and score the body's output against every token. Neither job obviously needs 640 learned floats per token. Our earlier released models did without a learned table. SHADOW-250M \cite{r25} is a 250M-parameter model trained on 30B tokens whose 131,072 tokens each carry a fixed 512-bit code and nothing else; with an attention and FFN body under 2 bits per weight it ships as a 60 MB file, vocabulary included, and generates about 400 tokens per second on a laptop CPU in 80 MB of RAM, at a held-out perplexity of 23.3 and PIQA 60, HellaSwag 34, ARC-Easy 41 (normalised accuracy). SHADOW-50M \cite{r26} is a 44M-parameter ternary model over a frozen 73,880-token table, 19.8 MB in total, about 2,000 tokens per second, with exact arithmetic circuits and a memory on disk; it scores below a normal model of its size on the usual tests and says so. Both worked, but neither was a controlled experiment: they changed many things at once, and neither was ever trained with a learned table for comparison. This paper is the controlled version. Same body, same data, same steps, same seed. Only the vocabulary changes.

\begin{table}[H]
\caption{\textbf{Where Gemma 3 270M's parameters are.} From the public configuration: 18 layers, width 640, FFN 2,048, 4 query and 1 KV head of size 256, vocabulary 262,144.}
\centering\small
\begin{tabular}{lll}
\toprule
\textbf{part} & \textbf{parameters} & \textbf{share} \\
\midrule
embedding table (tied to the head) & 167.8M & 62.6\% \\
feed-forward layers & 70.8M & 26.4\% \\
attention & 29.5M & 11.0\% \\
norms & 0.06M & 0.0\% \\
total & 268.1M &  \\
\bottomrule
\end{tabular}
\end{table}

\section{The frozen table}

The idea is simple enough to state in one paragraph. Take an existing embedding matrix E (we use Gemma 3 270M's own, 262,144 by 640). Centre it, take the top 512 principal directions W, and keep only the sign of each projection:

    code(t) = sign( W\textasciicircum{}T (E\_t - mu) )   in \{-1, +1\}\textasciicircum{}512

That is 512 bits per token, 16.8 MB for the whole vocabulary, and it never changes again. Going in, a token's code passes through a small trained linear map (512 to 640) into the body. Coming out, the body's final state passes through another small map (640 to 512) and is scored against every code by a dot product, divided by sqrt(512). The two maps are 0.65M parameters. A fixed per-token prior, the log unigram frequency of each token in the training data, is added to the scores; it costs 1 MB and has no trainable parameters. We call it the seal. Arm C lets this prior be trained from that starting point, which adds 262,144 trainable numbers.

The table keeps most of what the pretrained embedding knew. The 512 directions carry 92.3\% of E's variance. On WordSim-353, the codes' agreement correlates with human similarity ratings at 0.617, against 0.614 for the float embedding itself. Of each token's 10 nearest float neighbours, 67\% are among its 10 nearest code neighbours.

Because the table is a function of E, any vector in E's space can become a token after training: a new word gets code(mean of its pieces' rows), or code of any vector you choose. A learned table has no such rule. A new row in it is random until it is trained.

\section{Experiment}

\textbf{Body.} Gemma 3 270M's architecture exactly, built from the public configuration with random initialisation: 18 layers, width 640, gated GELU FFN of 2,048, 4 query and 1 key-value head of size 256 with QK-norm, five local (512-token sliding window) layers per global layer, RoPE base 10k local and 1M global, RMSNorm, no biases, no dropout. We trained from scratch in every arm.

\textbf{Arms.}

\textbf{Arm A.} The learned tied table, 262,144 by 640, with the body in bfloat16 autocast. This is the model as Google would train it, at our data scale.

\textbf{Arm B.} The frozen table and the fixed seal, same float body.

\textbf{Arm C.} The frozen table, every linear layer in the body replaced by BitNet b1.58's BitLinear (RMSNorm on the input, int8 absmax activations per token, absmean ternary weights, straight-through estimator) \cite{r2,r4}, attention arithmetic and KV cache in 16-bit as in BitNet, and the seal trained from its unigram start. A has 268.1M trainable parameters; B 101.0M; C 101.2M.

\textbf{Data and schedule.} 2.05B tokens of FineWeb-Edu (sample-10BT) \cite{r5} tokenised with Gemma's tokenizer, documents separated by the end-of-sequence token, 10M held-out validation tokens. Sequence length 1,024; 262,144 tokens per optimiser step; 7,820 steps. Every arm draws exactly the same batches in the same order from the same seed. A and B use AdamW with peak learning rate 6e-4, betas 0.9 and 0.95, weight decay 0.1, 1\% warm-up and cosine decay to 10\%. C uses the recipe BitNet's authors published for their smallest model \cite{r4}: peak 1.5e-3 decaying linearly, dropping to a 1e-3 peak at the midpoint, weight decay 0.1 in the first half and 0 in the second, 375 warm-up steps. We did not tune anything for C; we took their numbers. This means C differs from B in three ways (body, schedule, trained prior), which we return to in Section 6.

\textbf{Cost.} Each arm trained on one RTX 5090 in 14 to 16 hours, about 18 US dollars per arm at the rental price we paid. The whole study, including evaluation, cost under 75 dollars.

\textbf{Evaluation.} Validation loss on 512 held-out sequences of 1,024 tokens. LAMBADA, HellaSwag, ARC-Easy and PIQA with the lm-evaluation-harness \cite{r6}, zero-shot, full test sets. Every per-question result is saved, and every error bar in this paper is a 95\% bootstrap range over those questions (10,000 re-draws). Differences between arms are paired on the same questions.

\section{Results}

\begin{table}[H]
\caption{\textbf{Loss and zero-shot accuracy.} Accuracy in \%, 95\% range in brackets; loss in nats per token.}
\centering\small
\resizebox{\linewidth}{!}{\begin{tabular}{lllllll}
\toprule
 & \textbf{trainable} & \textbf{LAMBADA} & \textbf{HellaSwag} & \textbf{ARC-Easy} & \textbf{PIQA} & \textbf{val loss} \\
\midrule
A learned table, float & 268.1M & 21.6 (20.5-22.7) & 30.4 (29.4-31.2) & 45.9 (43.9-47.9) & 60.6 (58.3-62.8) & 3.155 \\
B frozen table, float & 101.0M & 23.2 (22.1-24.4) & 30.4 (29.4-31.2) & 42.5 (40.5-44.5) & 61.5 (59.2-63.8) & 3.358 \\
C frozen table, ternary & 101.2M & 22.5 (21.3-23.6) & 29.4 (28.6-30.3) & 41.0 (39.0-43.0) & 59.5 (57.2-61.7) & 3.439 \\
\bottomrule
\end{tabular}}
\end{table}

\begin{table}[H]
\caption{\textbf{Paired differences.} Points of accuracy, 95\% range; ``tied'' means the range includes zero.}
\centering\small
\resizebox{\linewidth}{!}{\begin{tabular}{llllll}
\toprule
 & \textbf{LAMBADA} & \textbf{HellaSwag} & \textbf{ARC-Easy} & \textbf{PIQA} & \textbf{loss} \\
\midrule
B minus A & +1.6 (+0.7 to +2.6) & 0.0 (-0.6 to +0.6) tied & -3.4 (-5.1 to -1.7) & +0.9 (-0.8 to +2.6) tied & +0.203 \\
C minus A & +0.9 (-0.1 to +1.8) tied & -0.9 (-1.6 to -0.3) & -4.9 (-6.7 to -3.1) & -1.1 (-2.8 to +0.6) tied & +0.284 \\
C minus B & -0.8 (-1.7 to +0.2) tied & -0.9 (-1.5 to -0.3) & -1.5 (-2.9 to +0.1) tied & -2.0 (-3.5 to -0.5) & +0.081 \\
\bottomrule
\end{tabular}}
\end{table}

\textbf{What the learned table buys.} In loss, 0.20 nats, which is about 23\% lower perplexity. On the benchmarks, much less. B is better than A on LAMBADA by 1.6 points, tied on HellaSwag and PIQA, and worse on ARC-Easy by 3.4 points. That is the whole return on 168M trained parameters: one benchmark.

\textbf{How the gap develops.} Figure 1 shows validation loss over training, checked every 200 steps. B starts ahead: at step 200 it is 0.23 nats below A, because its codes already carry Gemma's meanings and its prior already knows word frequencies. A overtakes at step 600, pulls ahead until about step 3,000, and from there the gap stays at 0.20 to the end. B's loss falls at every check. The learned table's advantage is not a head start that fades; it is a capacity that A grows into once it has seen enough data to train the table.

\textbf{Where the gap is.} Table 4 splits the validation loss by how often the target token appeared in training. The learned table wins in the middle: tokens seen between 100 and 100,000 times, where there are enough examples to train a row and the row then beats a fixed code. On the 161 most common tokens, which are 52\% of all text, the gap is 0.035 nats. On tokens seen 10 to 99 times, the frozen table wins by 2.3 nats (B) and 1.7 nats (C): the learned rows never got enough examples, and the codes arrived knowing the word.

\begin{table}[H]
\caption{\textbf{Validation loss by the target token's count in training.} 512 sequences; differences paired, 95\% range.}
\centering\small
\resizebox{\linewidth}{!}{\begin{tabular}{lllllll}
\toprule
\textbf{count in training} & \textbf{tokens in vocab} & \textbf{share of text} & \textbf{A} & \textbf{B} & \textbf{C} & \textbf{B minus A} \\
\midrule
10-99 & 41,004 & 0.04\% & 14.36 & 12.01 & 12.66 & -2.34 (-2.95 to -1.74) \\
100-999 & 43,577 & 0.9\% & 7.44 & 8.32 & 8.45 & +0.89 (+0.79 to +0.98) \\
1k-9.9k & 48,178 & 7.9\% & 5.71 & 6.51 & 6.60 & +0.80 (+0.78 to +0.83) \\
10k-99k & 12,382 & 16.8\% & 4.79 & 5.21 & 5.36 & +0.42 (+0.41 to +0.43) \\
100k-999k & 1,885 & 22.3\% & 3.87 & 4.07 & 4.18 & +0.20 (+0.19 to +0.20) \\
1M and more & 161 & 52.0\% & 1.85 & 1.88 & 1.93 & +0.035 (+0.032 to +0.037) \\
\bottomrule
\end{tabular}}
\end{table}

Of Gemma's 262,144 tokens, 52,782 never appear in our 2.05B training tokens and only 147,187 appear ten or more times. In A, the rows of the rest are random or barely moved from random. In B and C they still carry Gemma's codes.

\textbf{The ternary model.} C is about 40 MB at 1.58 bits per weight: 16.8 MB of table, about 20 MB of ternary body, 1.3 MB of 16-bit interface maps and 0.5 MB of prior. The file we actually run (next paragraph) packs the body at 2 bits per weight and keeps the maps, norms and scales in 32-bit, and is 46.2 MB. A in 16-bit is 543 MB as a GGUF file; B would be about 220 MB. C ties A on LAMBADA and PIQA, is 0.9 points behind on HellaSwag and 4.9 behind on ARC-Easy. Against B, which has the same table and a float body, C is within 0.8 to 2.0 points everywhere and 0.08 nats in loss. BitNet's two-stage schedule behaved as its authors describe: C's loss curve is S-shaped, and at the learning-rate drop at step 3,911 it fell 0.036 nats in 200 steps, 3.6 times B's fall over the same steps.

\textbf{On a laptop CPU.} We wrote a CPU kernel for C (released with the paper: one C file, AVX2, ternary matrix-vector products with int8 activations ported from our SHADOW-50M kernel \cite{r26}, and a readout that scores all 262,144 codes straight off the 1-bit table) and ran A through llama.cpp from its GGUF files, on the same machine, the same 8 threads, both with a 16-bit KV cache. Table 5 gives the result. The kernel reproduces the PyTorch model: on 64 validation tokens its argmax agrees at every position and its next-token loss is 4.495 against 4.483. There is no kernel for B, whose float body is the same as A's; it would sit between the two.

\begin{table}[H]
\caption{\textbf{Generation speed and memory on one laptop CPU.} Intel Xeon W-11955M, 8 threads, greedy decoding at a 128-token context; peak resident memory of the process while generating. Perplexity of the GGUF files on 300k validation tokens at context 512, measured with llama-perplexity, so that the 8-bit and 4-bit files can be read as the float model and not as a weaker one.}
\centering\small
\resizebox{\linewidth}{!}{\begin{tabular}{llllll}
\toprule
\textbf{model} & \textbf{runtime and format} & \textbf{file} & \textbf{peak memory} & \textbf{tokens per second} & \textbf{perplexity} \\
\midrule
A, learned table, float & llama.cpp, GGUF f16 & 543 MB & 723 MB & 62 & 23.86 (0.19) \\
A & llama.cpp, GGUF Q8\_0 & 292 MB & 471 MB & 116 & 23.86 (0.19) \\
A & llama.cpp, GGUF Q4\_K\_M & 253 MB & 446 MB & 130 & 23.97 (0.19) \\
C, frozen table, ternary & our kernel & 46 MB & 62 MB & 615 to 740 & 31.2 (from the loss in Table 2) \\
\bottomrule
\end{tabular}}
\end{table}

The 8-bit GGUF loses nothing (23.864 against 23.857, far inside the error), so the fair opponent for C is A at Q8\_0: C generates 5.3 to 6.4 times faster in 7.6 times less memory. A cannot get below 446 MB at any quantisation because llama.cpp, like every other quantiser, keeps the 262,144 by 640 table in 16-bit: 250 MB of the 4-bit file is the vocabulary. C's speed is set by memory traffic, not arithmetic: each generated token streams the 25 MB body and the 17 MB table through the processor, and this laptop's memory moves 38 GB/s, so its ceiling is about 860 tokens per second; a processor whose last-level cache holds the 46 MB would go faster. C's quality is lower than A's by the 0.28 nats of Table 2; that is the trade the table makes, and the benchmarks in Table 3 say how much of it is visible.

\begin{figure}[H]
\centering
\includegraphics[width=0.88\linewidth]{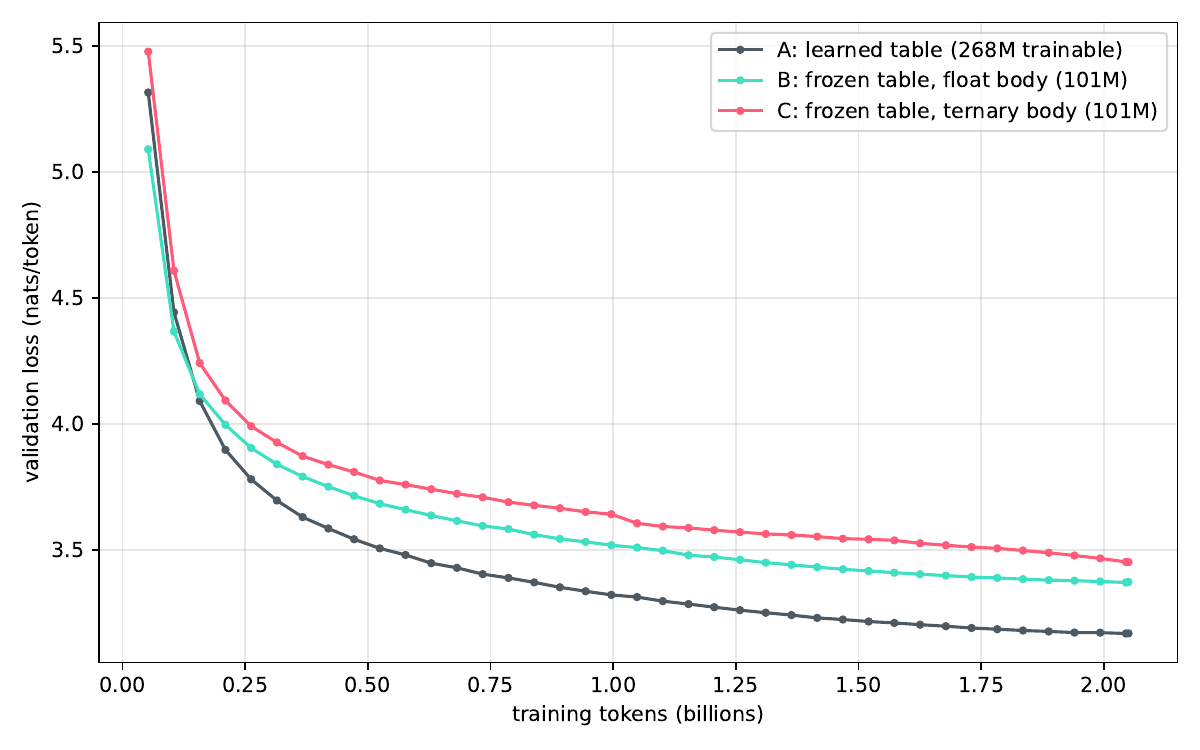}
\caption{Validation loss of the three arms over 2.05B training tokens, checked every 200 steps.}
\end{figure}

\begin{figure}[H]
\centering
\includegraphics[width=0.88\linewidth]{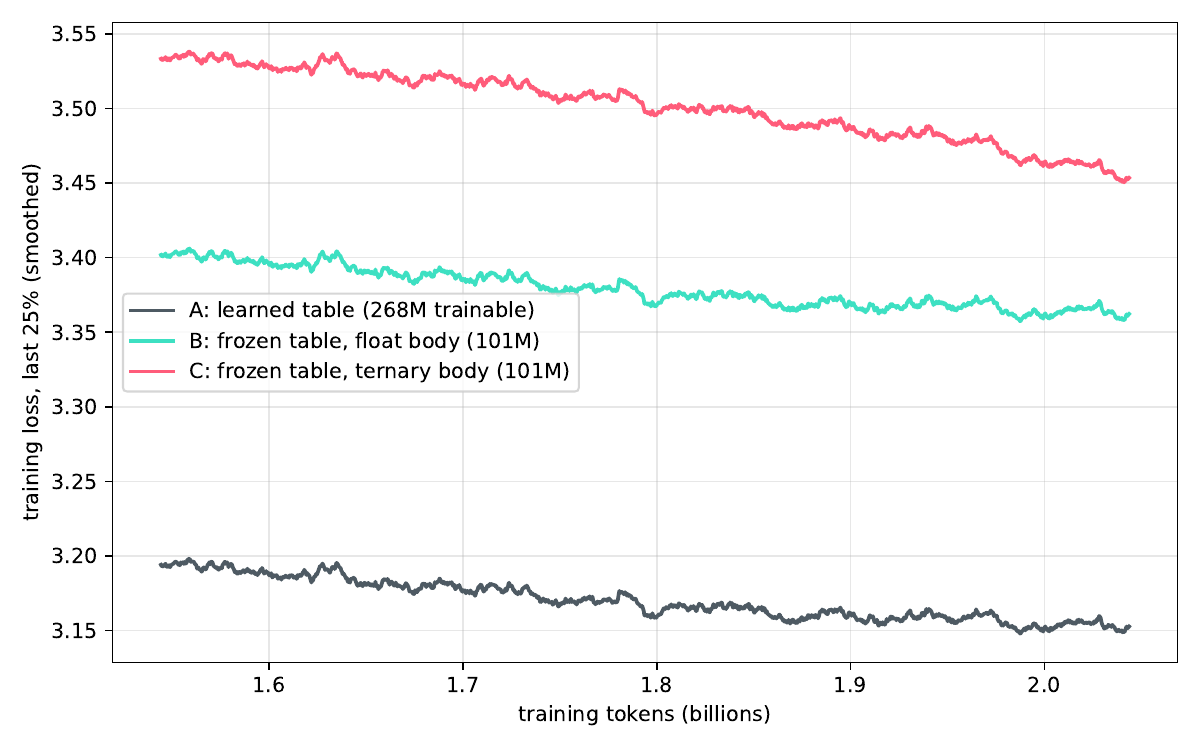}
\caption{The last quarter of training: smoothed training loss, tokens 1.55B to 2.05B.}
\end{figure}

\section{Adding to the vocabulary after training}

A learned table cannot take a new token without training: the new row is random. A frozen table can, because its codes come from a rule. We tested what that is worth, and we report the negatives with the positives.

\textbf{A domain vocabulary, no training.} We took PubMed articles \cite{r7}, found the 10,000 most frequent biomedical words that Gemma's tokenizer splits into two or more pieces (2.5 pieces on average), and added each as a single new token: for B and C, code(mean of its pieces' Gemma rows) with a prior set to the word's estimated training frequency, chosen on a validation split; for A, a new row at the mean norm of its learned rows, the fair untrained equivalent. On 300 held-out articles the new tokens replace 3.2\% of all tokens. Table 6 shows the cost.

\begin{table}[H]
\caption{\textbf{Quality after adding new tokens, no training.} Change in nats per character against the same model without the new tokens; 95\% ranges exclude zero in every cell.}
\centering\small
\begin{tabular}{lllll}
\toprule
\textbf{tokens added} & \textbf{tokens saved} & \textbf{A learned} & \textbf{B frozen} & \textbf{C frozen, ternary} \\
\midrule
2,000 & 1.9\% & +20.2\% & +2.0\% & +1.9\% \\
5,000 & 2.7\% & +32.6\% & +2.8\% & +2.7\% \\
10,000 & 3.2\% & +42.9\% & +3.3\% & +3.1\% \\
\bottomrule
\end{tabular}
\end{table}

The learned table collapses. The frozen table takes 10,000 new words at a 3\% cost and reads the text with 3\% fewer tokens. We tried several ways of setting the new tokens' priors and several ways of building their codes (mean of pieces, first piece, a code read off the model's own output before the word); none made the cost zero. The model's trained reading of the pieces is slightly better than any code we could construct for the whole word. So the honest claim is about possibility, not improvement: a frozen table accepts new rows and a learned table does not. The improvement case is the one our SHADOW-50M \cite{r26} met in practice, a table that could not spell 8,600 English pieces at all; Gemma's vocabulary leaves almost nothing of that kind to add (0.001\% of held-out tokens fall back to bytes).

\textbf{Tokens the model never trained on.} On text in 20 languages, B is better than A overall in Arabic, Bulgarian, Spanish, Russian and Vietnamese and worse in the other fifteen. Tokens that never occurred in training are penalised in B by the seal, which gives a count of zero the prior log(1/N). Setting those tokens' prior to the median of the seen tokens, with no training, makes B better than A on never-seen tokens in Thai, Arabic, Russian, French and Greek; A stays ahead in Hindi, Japanese and Chinese. The codes carry meaning for tokens the model never saw; the prior has to be told not to forbid them.

\textbf{What did not work.} Two tests we expected to favour the frozen table did not. Brand-new control tokens (\texttt{<tool\_call>}, \texttt{<think>}, \texttt{<sql>}, citation and field markers, five families, 24 prompt rotations each) are placed correctly from in-context examples by every arm: at five shots, B 88\%, C 75\%, A 73\%; at one shot A is highest, from rows that happened to score high with no examples. That test measures in-context learning, which every arm has, and the luck of the random row. Emoji prediction for tweets (SemEval-2018 task 2, 20 classes) is at chance for all three models, 2.6 to 5.7\% against 5\%; a model trained on 2B English tokens has no tweet-to-emoji knowledge for a code to unlock. Both results are released with the rest.

\section{Limitations}

One seed per arm. The error bars are over test questions, not over training runs; a second seed could move the loss gap by a few hundredths. Two billion tokens is Chinchilla-optimal for the 101M-parameter arms and under-trains A's table; it is also the budget we had. The learned table's advantage grows with data per token, so the loss gap at 20B tokens is likely larger than 0.20, and the benchmark picture may change with it.

C differs from B in three things, not one: the ternary body, BitNet's schedule, and the trained prior. We measured the prior's share in a separate run of B with the trained prior that we stopped at step 1,833 for budget: it was 0.003 to 0.022 nats better than B at every matched check. The schedule's share we did not isolate.

The frozen codes are derived from Gemma 3 270M's embedding, so they inherit whatever that embedding knows and does not know, and a release of the table carries Google's Gemma terms. The method needs some pretrained embedding to start from; an alternative is to train a small model with a learned table once and freeze its codes for everything after, which we have not tested here.

Adding vocabulary after training costs quality in every setting we tried where the tokenizer already covered the words. The gain is bounded by what the tokenizer is missing.

The CPU comparison is one laptop and two different code bases: our kernel for C against llama.cpp for A. Both are memory-bound, so the ratio should hold on other machines while the absolute numbers scale with memory bandwidth, but we measured only the one machine. B has no kernel. Our kernel's prefill is one token at a time, so we compare generation speed only.

\section{Related work}

Our own SHADOW-250M and SHADOW-50M \cite{r25,r26} are the origin of the frozen 512-bit table and of the CPU kernels used here; this paper supplies the controlled comparison those releases did not have. Bochkov \cite{r8,r9} trains transformers from scratch with frozen binary codes derived from token identity on the input and a trained float output head; the codes carry no meaning and the output table stays. Our codes come from a pretrained embedding, are shared by input and output, and leave no vocabulary matrix on either side. Press and Wolf \cite{r10} tie input and output embeddings; ALBERT \cite{r11} factorises the table into a small learned one and a projection; hash embeddings \cite{r12} share rows between tokens. All keep the table trainable and float. Meister et al. \cite{r13} initialise a per-token output bias to the log unigram distribution and train it, which is what our arm C does and arm B freezes. BitNet \cite{r1,r2,r3,r4} gives us the ternary body and the schedule; the BitNet papers report model memory excluding the embedding. GPTQ \cite{r14} and AWQ \cite{r15} quantise the body after training and leave the embedding in 16-bit. FineWeb-Edu \cite{r5} is the training data, Gemma 3 \cite{r16} the architecture and the source embedding, and the straight-through estimator \cite{r17} the training trick for ternary weights.

\section{What we release}

Everything the paper used, at the repository above: the training script with every setting; the frozen table, the unigram counts and the seal; the 2.05B training tokens and the validation tokens with checksums; all intermediate and final checkpoints of A, B and C (weights and optimiser state, every 800 steps) and the partial run of B with the trained prior; the per-step training logs; the evaluation scripts and every per-question benchmark result, so any number in this paper can be recomputed, and any other arm can be added to the same tables. Licence: non-commercial research. The table and the checkpoints of B and C contain codes derived from Gemma 3 270M and are released under Google's Gemma terms as well.

\section{What comes next}

A table that is a rule rather than a trained matrix does not have to stop at words. The next paper puts one frozen table under text, images, audio and video at once: every kind of input and output lands in the same 512-bit code space, so one small body reads a picture, hears a sound and writes a sentence with no learned embedding anywhere, and the whole thing still fits on a CPU. The teaser is all we will say here; the measurements come with the paper.

\section*{How to cite}

Sai Kiran Bathula. Embeddings Are Overpriced? A Frozen 16 MB Vocabulary for Small and Ternary Language Models. Preprint, 2026. doi.org/10.5281/zenodo.23382634. Code, data and weights: \url{huggingface.co/QLNI/embeddings-are-overpriced}.

\begin{small}\begin{verbatim}
@misc{bathula2026embeddings,
  title  = {Embeddings Are Overpriced? A Frozen 16 MB Vocabulary
            for Small and Ternary Language Models},
  author = {Bathula, Sai Kiran},
  year   = {2026},
  doi    = {10.5281/zenodo.23382634},
  note   = {Preprint. Code, data and weights:
            huggingface.co/QLNI/embeddings-are-overpriced}
}
\end{verbatim}\end{small}

\end{document}